\documentclass{article} 

\PassOptionsToPackage{table,xcdraw}{xcolor}

\usepackage{iclr2027_conference,times}

\usepackage{amsmath,amsfonts,bm}

\def\eqref#1{equation~\ref{#1}}

\def\1{\bm{1}}

\DeclareMathAlphabet{\mathsfit}{\encodingdefault}{\sfdefault}{m}{sl}
\SetMathAlphabet{\mathsfit}{bold}{\encodingdefault}{\sfdefault}{bx}{n}

\definecolor{iccvblue}{rgb}{0.21,0.49,0.74}
\usepackage[breaklinks,colorlinks,allcolors=iccvblue]{hyperref}

\usepackage{url}
\usepackage{amsmath}
\usepackage{amssymb}
\usepackage{booktabs}
\usepackage{multirow}
\usepackage{tcolorbox}
\definecolor{best}{HTML}{C8E6C9}
\definecolor{best2}{HTML}{BBDEFB}
\definecolor{grey}{HTML}{e0e0e0}

\usepackage{algorithm}
\usepackage{algorithmic}
\usepackage{pifont}
\usepackage{cuted} 
\usepackage{newfloat}
\usepackage{listings}
\usepackage{soul}
\usepackage{marvosym}
\usepackage{wrapfig}
\usepackage{caption}  
\usepackage{tocloft}
\usepackage{etoc}
\usepackage{enumitem}
\usepackage[T1]{fontenc} 
\usepackage{float}
\usepackage{graphicx}

\definecolor{deepblue}{RGB}{0,0,180}        
\definecolor{mygreen}{RGB}{0,150,0}         

\title{VepAgent: Bridging Causal-Transition via Tool-Augmented Reinforcement Learning for Video Event Prediction}

\author{
{Qiutong Chen}\textsuperscript{1}$^{*}$,
{Yuchan Guo}\textsuperscript{2}$^{*}$,
{Zhenlong Yuan}\textsuperscript{3}$^{*}$,
{Haobo Yang}\textsuperscript{4},
{Fangfang Lin}\textsuperscript{5},
{Xinyi Long}\textsuperscript{3},    \\
\textbf{Yin Wang}\textsuperscript{6},
\textbf{Zijian Song}\textsuperscript{3},
\textbf{Rui Lan}\textsuperscript{3},
\textbf{Shi Qiu}\textsuperscript{3},
\textbf{Boyuan Pan}\textsuperscript{3},
\textbf{Yang Luo}\textsuperscript{3},
\textbf{Yuyin Zhou}\textsuperscript{7}\textsuperscript{\textsection} \\ \\
\textsuperscript{1} Nankai University,
\textsuperscript{2} Carnegie Mellon University,
\textsuperscript{3} Xiaohongshu Inc.,
\textsuperscript{4} Columbia University,   \\
\textsuperscript{5} Santa Clara University,
\textsuperscript{6} New York University, 
\textsuperscript{7} University of California, Santa Cruz \\ \\
\textsuperscript{*}~\textit{Equal contribution}\quad
\textsuperscript{\textsection}~\textit{Corresponding Author} \\ \\
{\footnotesize
  Code: \url{https://anonymous.4open.science/r/VepAgent-42F8}
}
}

\begin{document}

\maketitle

\begin{abstract}
Multimodal Large Language Models (MLLMs) have demonstrated remarkable potential in video understanding, yet their reliance on retrospective summarization and text-centric priors often limits their ability to bridge unobserved causal transitions when applied to Video Event Prediction (VEP). To address this, we propose \textbf{VepAgent}, an agentic framework that integrates causal-transition reasoning with tool-augmented reinforcement learning (RL) for robust VEP. Unlike prior methods that passively project future trajectories from historical dependencies, our approach explicitly models the logical progression from terminal observed states to future events. Specifically, we first construct \emph{futurebench-4K}, a high-quality chain-of-thought dataset for supervised fine-tuning (SFT) that effectively bridges the causal-logic gap by structuring the deduction of unobserved intermediate states. Subsequently, we develop a diagnostic tool library integrating state tracking, frame retrieval, and region magnification, enabling the agent to dynamically augment reasoning with external tools to recover missing spatio-temporal evidence and resolve visual ambiguities during inference. Moreover, we propose a composite reward mechanism that jointly optimizes prediction accuracy, causal coherence, and reliable prior, compelling the agent to rely on genuine visual grounding rather than superficial textual similarities. Extensive evaluations on FutureBench and NEPBench datasets demonstrate that our method achieves state-of-the-art performance, significantly outperforming larger MLLMs and validating the empirical effectiveness of our agentic, future-oriented reasoning paradigm.
\end{abstract}


\section{Introduction} Video Event Prediction (VEP) aims to anticipate plausible future events based on partially observed visual evidence~\citep{nako2025navigating, schmidt2025video}. This forward-looking capability is foundational for next-generation applications in areas including proactive robotics~\citep{abbate2024self}, autonomous driving~\citep{hu2020probabilistic}, intelligent surveillance~\citep{karim2022dynamic}, etc. Unlike standard video understanding tasks that primarily require descriptive summarization of observed content, VEP demands a deeper causal comprehension of temporal dynamics to future trajectories. A central challenge is developing agents that can bridge observed reality with unobserved futures under incomplete visual evidence.

Recent breakthroughs in Multimodal Large Language Models (MLLMs)~\citep{jain2024vcoder, wu2024visual, li2024mmedagent} have advanced both video perception and reasoning. By leveraging vast world knowledge and grounding it in visual evidence, MLLMs~\citep{khattak2025good} interpret nuanced visual details and generate coherent narratives from video inputs with high fidelity. Models like the Qwen-VL series~\citep{bai2025qwen2, bai2025qwen3} have demonstrated strong performance across a spectrum of challenging tasks, including video dialogue, temporal grounding, and few-shot learning.

Despite their capabilities, applying MLLMs to VEP still exposes three fundamental limitations. \ding{182} \textbf{Causal-Logic Gap.} While MLLMs can successfully reconstruct historical event chains, this retrospective summarization does not inherently confer the ability to extrapolate future trajectories. There exists an unobserved intermediate state transition between the terminal frame and the future event that requires structured causal bridging, rather than merely relying on historical dependencies to project outcomes. \ding{183} \textbf{Language-Prior Shortcut.} Instead of faithfully grounding reasoning in video content, models frequently exploit linguistic priors by selecting the answer option that shares the highest textual similarity with the query. This over-reliance on surface-level lexical patterns and statistical co-occurrences often overrides actual visual states, leading to visually unsupported predictions. \ding{184} \textbf{Visual-Evidence Omission.} Real-world videos frequently suffer from partial occlusions, rapid motion, or suboptimal frame sampling, truncating critical spatio-temporal cues for state verification. Without mechanisms to dynamically recover missing transient actions or fine-grained spatial details, MLLMs are forced to resort to biased estimations in ambiguous scenarios.

\begin{figure}[t]
    \centering
    \includegraphics[width=\textwidth]{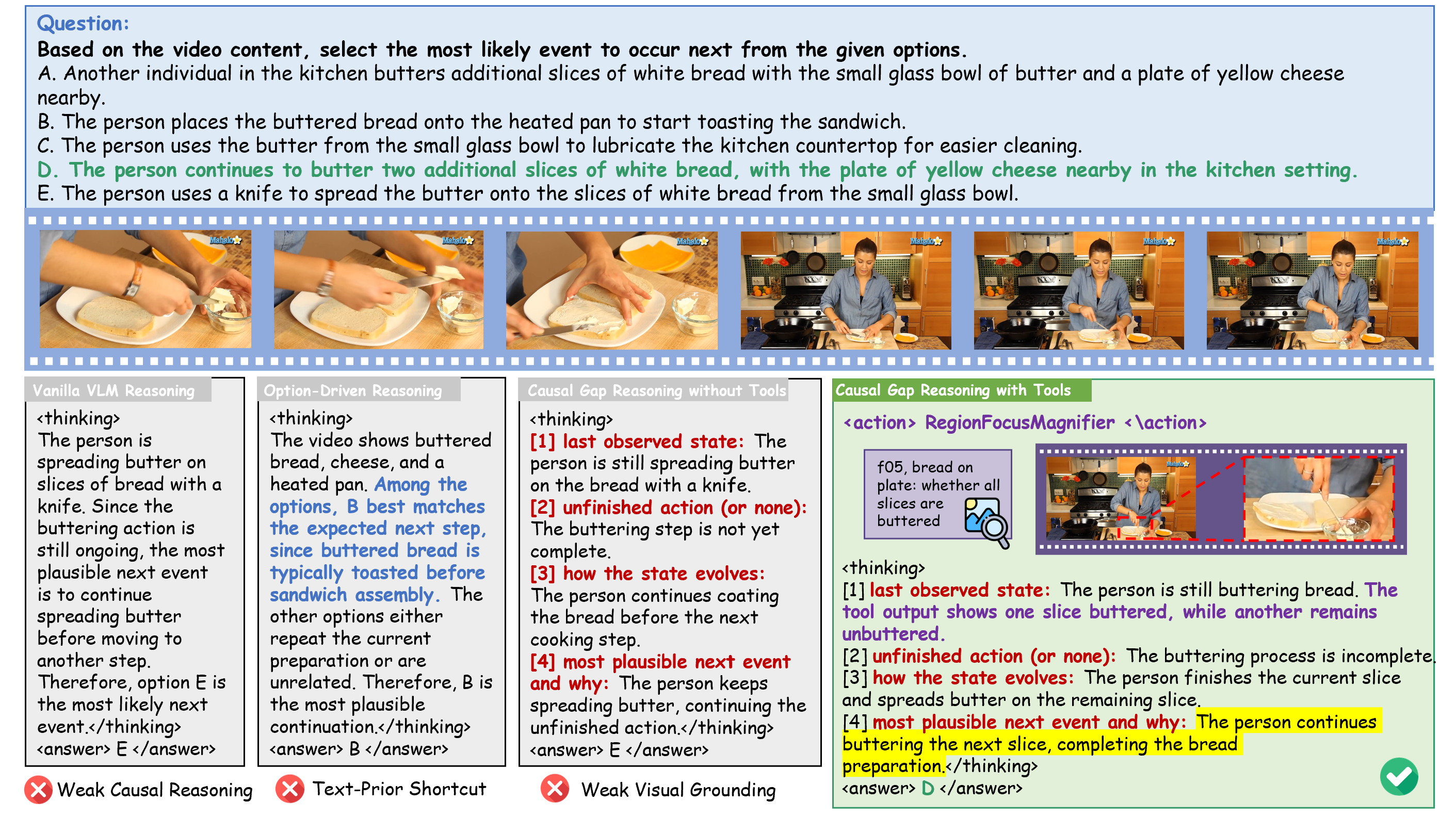}
    \caption{
    \textbf{Key insight of VepAgent.}
    (a) Vanilla VLM reasoning weakly models the future transition and predicts a coarse continuation of the observed action.
    (b) Option-driven reasoning is distracted by textual priors and prematurely predicts the cooking stage.
    (c) Causal-gap reasoning captures the ongoing buttering process, but misses the fine-grained visual state and selects the wrong continuation.
    (d) With tool-augmented visual evidence, VepAgent identifies the remaining unbuttered slice, correctly reasons over the unfinished action, and predicts the next event.
    }
    \vspace{-0.1in}
    \label{fig:key_insight}
\end{figure}

To address these challenges, we argue that an effective VEP agent requires three synergistic capabilities. \ding{182} \textbf{Causal-Transition Reasoning}, which explicitly models the logical trajectory from the terminal observed state, through the ongoing action, to the future event, enabling the model to construct physically plausible predictions rather than relying on historical summaries. \ding{183} \textbf{Anti-Prior Reinforcement}, which penalizes reasoning shortcuts that exploit superficial textual similarities between queries and options, thereby encouraging predictions based on actual video states. \ding{184} \textbf{Tool-Augmented Perception}, which dynamically integrates a multimodal tool library to recover omitted visual evidence and magnify fine-grained spatio-temporal details, transforming the static observer into an active perceiver capable of capturing fleeting state transitions. Together, these capabilities reformulate VEP from passive future guessing into an active process of causal analysis, visual verification, and temporal grounding.

Therefore, we introduce \textbf{VepAgent}, a framework that unifies causal-transition reasoning with \textbf{tool-augmented reinforcement learning (RL)} and an anti-prior reward for video event prediction. The \textbf{\emph{futurebench-4K}} chain-of-thought (CoT) dataset for \textbf{supervised fine-tuning (SFT)} structures the logical progression from observed states and ongoing actions to future events. It teaches the model to identify unfinished dynamics and physically plausible continuations rather than merely summarizing the past, bridging the causal-logic gap.

To overcome language-prior shortcuts and resolve visual-evidence omission, we further develop a tool-augmented framework tailored to group relative policy optimization (GRPO) in the RL stage. During inference, the agent dynamically interacts with a diagnostic tool library that tracks state transitions, retrieves frame evidence, magnifies local regions, and extracts fine-grained visual facts. A composite reward jointly optimizes prediction accuracy, causal-gap bridging, and an explicit Anti-Prior penalty, guiding the agent to learn not only \emph{how} to use external tools, but also \emph{when} and \emph{why} additional evidence should be acquired for grounded future predictions.
Our contributions are: 
\begin{itemize}[leftmargin=*,nosep] 
\item \textbf{Tool-Augmented Agentic RL.} We introduce a tool-augmented agentic RL framework that unifies causal-transition reasoning with dynamic evidence acquisition. Structured SFT and a Causal Gap Reward enable VEP through dynamic, future-oriented reasoning rather than retrospective summarization. 
\item \textbf{Composite Anti-Prior Reward.} We design a composite reward with an Anti-Prior term that penalizes language-shortcut behaviors and encourages reliance on visual evidence rather than superficial textual similarities. 
\item \textbf{Spatiotemporal Multi-Tool Integration.} We build a multimodal tool library integrating state tracking, focused frame retrieval, region magnification, and visual detail extraction, dynamically augmenting reasoning to resolve fine-grained detail ambiguity and key-frame omission. 
\item \textbf{Empirical Effectiveness.} On FutureBench and NEPBench, VepAgent (4B) raises average accuracy by 14.58 and 22.70 percentage points over the prior best MLLM baselines, from 66.86\% to 81.44\% and from 47.50\% to 70.20\%, respectively, under the same evaluation protocols. 
\end{itemize}

\section{Related Work}
\textbf{Video Event Prediction.}
Previous VEP methods primarily originated from script event induction in NLP~\citep{granroth2016mcnc} and early action anticipation~\citep{girdhar2021antvc}, relying on structured datasets like VidSitu~\citep{sadhu2021vidsitu} and EPIC-Kitchens~\citep{damen2018epic} to model dynamic event evolution. To capture complex semantic hierarchies, recent works like VidEvent~\citep{liang2025videvent} and EventFormer~\citep{su2025eventformer} introduced large-scale structured benchmarks and node-graph attention mechanisms for action-centric prediction. 
Earlier work such as VLEP~\citep{lei2020vlep} formulated next-event prediction from partially observed videos, while more recent benchmarks like FutureBench~\citep{wang2026fostering} evaluate future-event reasoning in modern MLLMs.
However, native MLLMs~\citep{su2026video, chen2025seedbenchr1} still suffer from insufficient visual utilization and language-prior shortcuts when extrapolating future trajectories. \textbf{VepAgent} combines MLLM backbones with tool-augmented RL to target these failure modes in video event prediction.

\textbf{Multimodal LLMs Reasoning.}
Recent advances in Large Language Models (LLMs) have demonstrated that RL-based post-training can enhance reasoning capabilities, as exemplified by Qwen-VL~\citep{bai2025qwen3}, OpenAI-o1~\citep{jaech2024openaio1}, DeepSeek-R1~\citep{deepseekr1}, etc. These paradigms have been extended to MLLMs for tasks like mathematical VQA~\citep{peng2025lmmr1}, image segmentation~\citep{liu2025seg}, and video understanding~\citep{feng2025videor1}. However, existing methods struggle with language-prior shortcuts~\citep{chen2025longvilar1, feng2026onethinker}. To address this, we propose causal-transition reasoning via tool-based RL for visual grounding.

\textbf{Tool-Augmented Agentic System.}
Early works like FAST~\citep{sun2025fast} and MVoT~\citep{li2025MVoT} introduce visual evidence into reasoning, forming multimodal CoT for image tasks. LLaVA-Plus~\citep{liu2023llavapluslearningusetools} pioneered training strategies for tool use, while VPD~\citep{hu2024visual} leveraged program-derived data to transfer tool skills. 
Recent methods like TACO~\citep{ma2024taco} and PyVision~\citep{zhao2025pyvision} extend tool use with RL, while VITAL~\citep{zhang2025thinking} further enables dynamic, on-demand visual evidence acquisition for video reasoning. Building on this line of work, VepAgent contributes VEP-specific causal-transition reasoning, with diagnostic tools designed to recover missing temporal transitions and fine-grained visual evidence.

\section{Methodology}

\begin{figure}[t]
    \centering
    \includegraphics[width=\textwidth]{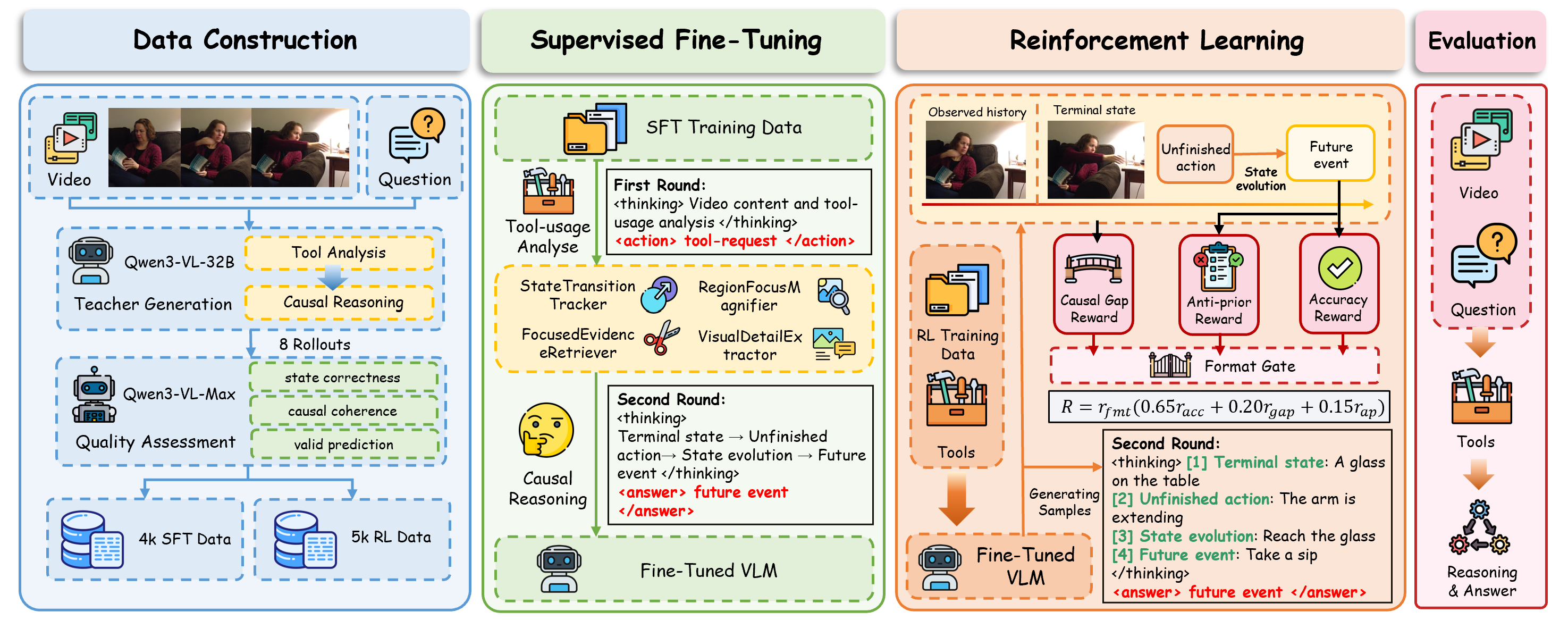}
    \caption{
    \textbf{Pipeline of VepAgent.}
    (i) Construct tool-augmented causal-transition CoT data by explicitly modeling the progression from terminal observed states to future events.
    (ii) Supervised fine-tuning initializes tool-use and causal-gap reasoning behaviors.
    (iii) GRPO optimizes the policy with accuracy, causal-gap, and anti-prior rewards.
    (iv) The trained agent dynamically integrates diagnostic tools for grounded future-event prediction.
    }
    \vspace{-0.1in}
    \label{pipeline}
\end{figure}

\textbf{Overview.} We present VepAgent, a framework that integrates causal-transition reasoning with tool-augmented RL and an anti-prior reward for video event prediction, as shown in Fig.~\ref{pipeline}. We next detail the problem formulation and workflow, training-data construction, SFT, and GRPO-based RL.



\subsection{Problem Formulation \& Workflow Design}
\label{Problem}
The goal in VEP is to anticipate the most plausible future event based on partially observed visual evidence. The input is a partially observed video $V$, a query $Q$, and a set of candidate future events (options) $\mathcal{O} = \{o_k\}_{k=1}^K$, while the output is the predicted future event $y \in \mathcal{O}$. 
Unlike standard video understanding that relies on retrospective summarization, VEP demands forward-looking causal reasoning. Formally, the model must accurately infer the terminal observed state $S_{\text{final}}$ and identify the ongoing, unfinished action $A_{\text{ongoing}}$ from $V$, and then logically bridge the causal-logic gap, which represents the unobserved intermediate state transition, to project the next physically plausible state.

To achieve this goal and resolve the visual-evidence omission during inference, we design the following two-stage agentic workflow to integrate domain-specific diagnostic tools:


\textbf{Stage 1: Observation \& Tool Selection.}
Given a partially observed video $V$ and query $Q$, our model first forms an initial estimate of the terminal observed state $S_{\text{final}}$ and the ongoing action $A_{\text{ongoing}}$. Concurrently, it constructs a causal hypothesis of the future trajectory. For instances where critical state transitions are ambiguous, occluded, or missed due to uniform frame sampling, our agentic framework formulates a plan by selecting an appropriate subset of tools from a diagnostic tool library $\mathcal{T}$. This library comprises a State Transition Tracker ($\mathcal{T}_{\text{s}}$), a Focused Evidence Retriever ($\mathcal{T}_{\text{f}}$), a Region Focus Magnifier ($\mathcal{T}_{\text{r}}$), and a Visual Detail Extractor ($\mathcal{T}_{\text{v}}$).
The policy network $\pi_\theta$ performs a single tool-selection step to determine a subset $C \subseteq \mathcal{T}$ based on the initial visual parsing and the specific causal gap being evaluated.


\textbf{Stage 2: Causal Bridging \& Tool Augmentation.}
The second stage executes the selected tools once and integrates their returned evidence into the reasoning context before producing the final prediction.
The tool outputs are not merely concatenated; they enhance the agent's spatio-temporal understanding: $\mathcal{T}_{\text{s}}$ clarifies action sequences, $\mathcal{T}_{\text{f}}$ recovers transient actions, and $\mathcal{T}_{\text{r}}$/$\mathcal{T}_{\text{v}}$ resolve fine-grained ambiguities (e.g., hand-object contact).
The final prediction $y$ is then inferred by conditioning the model $\pi_\theta$ on these augmented inputs, defined by:
\begin{equation}
p(y | V \oplus V_{\text{aug}}, Q \oplus E_{\text{aug}}) = \pi_\theta(V \oplus V_{\text{aug}}, Q \oplus E_{\text{aug}}),
\end{equation}
where $\oplus$ denotes modality-specific augmentation, $V_{\text{aug}}$ and $E_{\text{aug}}$ respectively represent augmented visual and textual evidence, while $\pi_\theta$ is the policy network of MLLM parameterized by weights $\theta$.

\begin{figure}[t]
    \centering
    \includegraphics[width=\textwidth]{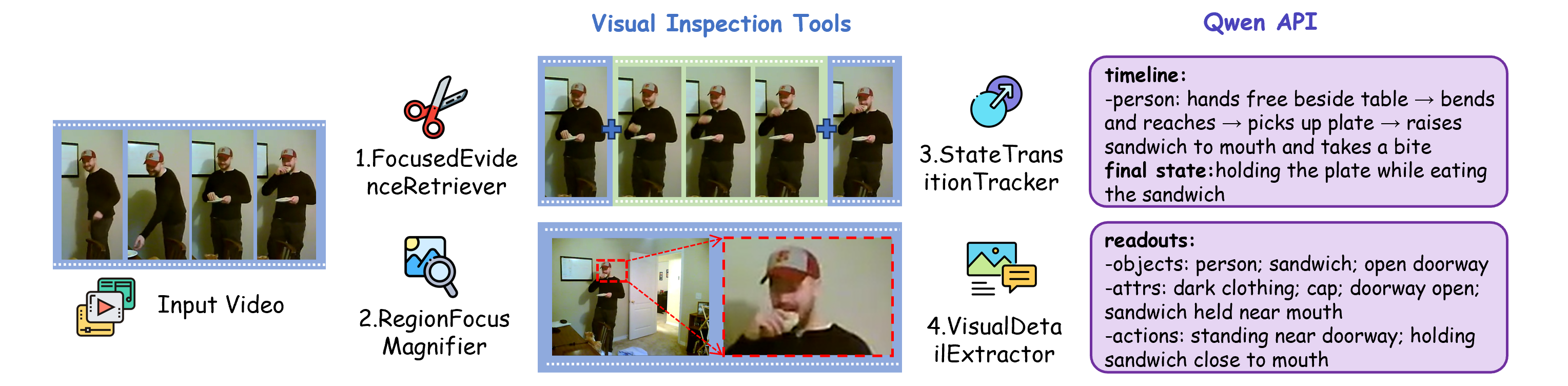}
    \caption{
    \textbf{Diagnostic tool library of VepAgent.}
    FRE retrieves dense visual evidence from selected temporal windows, while RFM
    crops and magnifies task-relevant spatial regions.
    STT summarizes observed state trajectories and the terminal state, whereas VDE
    extracts fine-grained visible facts as textual evidence.
    All tools operate exclusively on the observed portion of the video.
    }
    \vspace{-0.1in}
    \label{tool}
\end{figure}

\textbf{Tool Library.}
As shown in Fig.~\ref{tool}, we integrate four diagnostic tools for causal-transition reasoning.
\ding{182} \textbf{State Transition Tracker}: to resolve ambiguities in action order and state changes, this tool re-analyzes the video and structures the observed event chain and final state into a clear textual sequence.
\ding{183} \textbf{Focused Evidence Retriever}: to address the omission of actions caused by uniform sampling, this tool performs dense frame extraction within a specified temporal window.
\ding{184} \textbf{Region Focus Magnifier}: to resolve local detail ambiguity (e.g., whether a hand has released an object), this tool crops and magnifies key spatial regions for focused inspection.
\ding{185} \textbf{Visual Detail Extractor}: to mitigate overlooked fine-grained facts, this tool translates specific visual details (e.g., spatial relations) from the final frames into explicit textual descriptions, anchoring the causal reasoning in verifiable visual evidence. Details are provided in Appendix~\ref{appx:tool_library}.

\subsection{Training Data Construction}
\label{sec:Training}

A fundamental challenge when applying existing MLLMs to VEP is the absence of causal-transition reasoning chains: standard pipelines provide retrospective summarization of observed video content but lack intermediate reasoning steps---\textbf{what} the terminal observed state is, \textbf{which} action is ongoing but unfinished, and \textbf{how} the unobserved causal gap is bridged to project the future event. We therefore synthesize structured reasoning chains to construct \emph{futurebench-4K}, providing explicit examples of these steps.

\textbf{Data Collection.} Our data generation begins by leveraging the Qwen3-VL-32B~\citep{bai2025qwen3} to explore the base VEP training datasets. For each partially observed video, we perform 8 independent policy rollouts to generate diverse causal reasoning pathways. An instance is solvable if at least one rollout matches the ground-truth under the same matching protocol used for reward computation, ensuring the prediction is derived from actual visual states rather than language-prior shortcuts.

\textbf{Data Assessment.} To ensure the quality and reliability of the synthesized data, we employ the Qwen-VL-Max model as an expert validator to review and score each reasoning chain. Chains containing factual errors regarding observed states, logical inconsistencies in bridging the causal gap, or language-prior shortcut behaviors are discarded. This validation yields approximately 4,000 reasoning chains for SFT. The remaining successful trajectories, confirmed as valid but not selected for SFT, are used as the training data for the RL stage. To prevent data leakage, \emph{futurebench-4K} and all RL trajectories are constructed exclusively from the VEP training splits.

\textbf{Generating Pipeline.} Our generating pipeline implements a three-step causal-transition process. First, the agent identifies the terminal observed state $S_{\text{final}}$ and the ongoing, unfinished action $A_{\text{ongoing}}$. Second, if critical transitions are ambiguous, it conditionally introduces diagnostic tools (e.g., State Transition Tracker or Focused Evidence Retriever) to recover missing temporal evidence. Finally, the agent bridges the \textit{Causal Gap} by deducing how the unfinished action progresses to the next physically plausible state and the future-event prediction.

\subsection{Agentic Supervised Fine-Tuning}
\label{sec:SFT}
\textbf{After constructing the training data}, we introduce a \textbf{cold-start phase} that prioritizes supervised fine-tuning (SFT) before applying reinforcement learning (RL). 
Inspired by R1-Zero~\citep{deepseekr1}, we initially attempt direct RL optimization to train our method. However, preliminary experiments reveal a progressive decline in the frequency of tool invocation during policy rollouts. 
This behavior likely arises from a distributional discrepancy in target domain between tool-enhanced visual features and model's pretraining data. Therefore, we opt to perform SFT before RL.


Specifically, we formalize each training instance as $\mathcal{W} = (\mathcal{X}, \mathcal{Y}, \mathcal{Z}, \mathcal{O})$, where $\mathcal{X}$ denotes the input modality, $\mathcal{Y}$ represents task instructions, $\mathcal{Z}$ captures the structured reasoning process, and $\mathcal{O}$ is the final target output. During SFT, the reasoning process and final answer are concatenated into a complete assistant response $\tilde{\mathcal{Z}}=[\mathcal{Z};\mathcal{O}]=\{\tilde{z}_1,\dots,\tilde{z}_L\}$. We optimize the model with the standard autoregressive negative log-likelihood over the entire assistant response:
\begin{equation}
\mathcal{L}_{\text{SFT}} = -\mathbb{E}_{\mathcal{W} \sim \mathcal{D}} \left[ \sum_{t=1}^{L} \log p_{\theta}(\tilde{z}_t \mid \mathcal{X}, \mathcal{Y}, \tilde{z}_{<t}) \right],
\end{equation}
where $p_{\theta}(\cdot)$ denotes the conditional probability distribution. Thus, both the structured reasoning tokens and the final \texttt{<answer>} output are jointly optimized during SFT.

\subsection{Agentic Reinforcement Learning}
\label{sec:RL}
\textbf{Group Relative Policy Optimization (GRPO).}
We adopt the GRPO algorithm~\citep{shao2024deepseekmath} for policy optimization. 
GRPO employs a groupwise comparison framework to evaluate candidate responses.
Specifically, for each query $ q $ paired with its ground-truth solution $ a $ from dataset $ D $, the algorithm generates a set of rollout trajectories $ \{o_1, o_2, \dots, o_G\} $ based on the previous policy $ \pi_{\theta_{\text{old}}} $. The policy $ \pi_\theta $ is then refined through the optimization of this objective function:  
\begin{equation}
\begin{aligned}
    &\mathcal{L}_{GRPO}(\theta) = -\mathbb{E}_{q \sim P(Q), \{o_i\}_{i=1}^{G} \sim \pi_{\theta_{old}}(O|q)} \Bigg[ \frac{1}{G} \sum_{i=1}^G \\
    &\quad \Bigg( \min\left( \frac{\pi_\theta(o_i|q)}{\pi_{\theta_{old}}(o_i|q)} A_i, \text{clip}\left(\frac{\pi_\theta(o_i|q)}{\pi_{\theta_{old}}(o_i|q)}, 1-\epsilon, 1+\epsilon\right) A_i \right) - \beta \mathbb{D}_{KL}(\pi_\theta \| \pi_{ref}) \Bigg) \Bigg],
\end{aligned}
\end{equation}
\begin{equation}
\mathbb{D}_{KL}(\pi_\theta||\pi_{ref}) = \frac {\pi_{ref}(o_i|q)} {\pi_\theta(o_i|q)} - \log \frac {\pi_{ref}(o_i|q)} {\pi_\theta(o_i|q)} - 1,
\end{equation}
where $ \beta $ is adopted to balance the trade-off between exploration and stability during optimization. Then the advantage estimator $ A_{i} $ is calculated using normalized rewards from the trajectory group:  
\begin{equation}
A_{i} = \frac{r_i - \text{mean}(\{r_1, r_2, \dots, r_G\})}{\text{std}(\{r_1, r_2, \dots, r_G\})}.
\end{equation}
Each trajectory $ o_i $ receives a composite reward through rule-based verification designed as follows.

\textbf{Reward Design.} Effective reward functions should balance prediction accuracy, causal-logic bridging, and robustness against language-prior shortcuts. To this end, we design a composite reward that integrates three components: an accuracy reward $R_{\text{acc}}$, a causal gap reward $R_{\text{cau}}$, and an anti-prior reward $R_{\text{ant}}$, all gated by a format reward $R_{\text{format}}$. The format reward serves as a hard gate to ensure structural coherence, where $R_{\text{format}}(\tau)=1$ if the trajectory $\tau$ strictly follows the required reasoning format (e.g., containing valid \textit{<thinking>} and \textit{<answer>} tags), and $0$ otherwise. The accuracy reward $R_{\text{acc}}$ is a binary metric quantifying the correctness of the final prediction, where $R_{\text{acc}}(\tau)=1$ if the predicted future event semantically matches the ground-truth event, and $0$ otherwise.


Differently, the causal gap reward $R_{\text{cau}}$ quantifies whether the
model's reasoning genuinely bridges the unobserved intermediate state transition
rather than directly guessing the future. Specifically, it evaluates the
consistency of the predicted terminal state and causal transition with reference
reasoning, while assigning greater emphasis to the transition toward the future
event. The resulting process reward is further conditioned on prediction
correctness to prevent incorrect predictions from receiving excessive credit.

Furthermore, the anti-prior reward $R_{\text{ant}}$ is designed to mitigate the language-prior shortcut, where models merely select the option with the highest textual similarity to the query $Q$. 
For a given query and the correct option among $K$ candidates, we compute the textual similarity rank $k$ of the correct option with respect to $Q$ using sentence-embedding cosine similarity.
If the model predicts correctly, we assign a higher reward when the correct option is less textually guessable (i.e., a larger rank $k$). The anti-prior reward is formally defined as:
\begin{equation}
R_{\text{ant}}(\tau) = 
\begin{cases}
\frac{k - 1}{K - 1}, & \text{if } R_{\text{acc}}(\tau) = 1 \text{ and } K > 1 \\
0, & \text{otherwise}
\end{cases}
\end{equation}
This continuous formulation yields $0$ for the most obvious textual guess and $1$ for the least obvious, incentivizing the agent to rely on visual evidence rather than superficial textual priors. Ultimately, final reward for a trajectory $\tau$ is computed by combining these components, defined as:
\begin{equation}
R(\tau) = R_{\text{format}}(\tau) \cdot \left( 0.65 R_{\text{acc}}(\tau) + 0.20 R_{\text{cau}}(\tau) + 0.15 R_{\text{ant}}(\tau) \right).
\label{eq:total_reward}
\end{equation}
The composite reward balances prediction accuracy with causal coherence and an anti-prior penalty, encouraging visually grounded predictions.

\section{Experiment}
\textbf{Dataset and Metrics}.
We evaluate our method on two public VEP benchmarks: FutureBench~\citep{wang2026fostering} and NEPBench~\citep{lei2020vlep}. 
\textbf{FutureBench} measures the overall event prediction accuracy across different reasoning depths. Following prior work, we report accuracy on 1-Hop, 2-Hop, and 3-Hop predictions, as well as Interp. (Interpolation) and the overall AVG. 
\textbf{NEPBench} provides a comprehensive evaluation of Next Event Prediction capabilities across diverse real-world scenarios, containing subsets derived from Charades, VidSitu, and YouCook2. We report the Overall average accuracy and the individual accuracy on each subset. For both datasets, a prediction is considered correct only if it semantically matches the ground-truth future event.

\begin{table*}[t]
\centering
\caption{Event Prediction Acc. (\%) on FutureBench. \colorbox{best}{best} and \colorbox{best2}{second best} results are highlighted.}
\vspace{-0.5em}
\label{tab:futurebench}
\resizebox{0.75\textwidth}{!}{
\begin{tabular}{l|c|ccccc}
\toprule
\multicolumn{1}{c|}{Method} & \multicolumn{1}{c|}{Frames} & \multicolumn{1}{c}{1-Hop} & \multicolumn{1}{c}{2-Hop} & \multicolumn{1}{c}{3-Hop} & \multicolumn{1}{c}{Interp.} & \multicolumn{1}{c}{AVG} \\ \midrule
GLM-4.1V-9B          & 32 & 29.90 & 41.90 & 52.20 & 47.30 & 44.38 \\
LLaVA-NeXT-Video     & 32 & 48.80 & 49.30 & 40.00 & 44.40 & 45.24 \\
Kimi-VL-A3B          & 32 & 44.30 & 42.80 & 51.30 & 51.90 & 48.87 \\
MiMo-VL-7B           & 32 & 59.00 & 59.60 & 50.50 & 43.80 & 50.45 \\
InternVL3-8B         & 32 & 54.30 & 58.00 & 63.20 & 54.40 & 56.72 \\
Qwen2.5-VL-32B       & 32 & \cellcolor{best2}66.50 & 62.70 & 63.20 & 55.20 & 59.94 \\
Qwen3-VL-30B-A3B     & 32 & 65.30 & \cellcolor{best2}70.50 & \cellcolor{best2}76.10 & \cellcolor{best2}62.20 & \cellcolor{best2}66.86 \\
Qwen3-VL-4B & 32 & 54.91 & 57.51 & 59.20 & 61.15 & 59.09 \\
\midrule
\textbf{VepAgent-4B (Ours)}   & \textbf{32} & \cellcolor{best}\textbf{82.08} & \cellcolor{best}\textbf{88.60} & \cellcolor{best}\textbf{80.10} & \cellcolor{best}\textbf{78.94} & \cellcolor{best}\textbf{81.44} \\
\bottomrule
\end{tabular}
}
\vspace{-1em}
\end{table*}

\begin{table}[t]
\centering
\caption{Event Prediction Acc. (\%) on NEPBench. \colorbox{best}{best} and \colorbox{best2}{second best} results are highlighted.}
\vspace{-0.5em}
\label{tab:nepbench}
\resizebox{0.65\textwidth}{!}{
\begin{tabular}{l|cccc}
\toprule
\multicolumn{1}{c|}{Method} & \multicolumn{1}{c}{Overall} & \multicolumn{1}{c}{Charades} & \multicolumn{1}{c}{VidSitu} & \multicolumn{1}{c}{YouCook2} \\ \midrule
GLM-4.1V-9B          & 43.73 & \cellcolor{best2}45.98 & 28.18 & \cellcolor{best2}64.48 \\
InternVL3-8B         & 43.60 & 37.01 & 40.55 & 52.26 \\
Kimi-VL-A3B          & 41.20 & 28.66 & \cellcolor{best2}53.06 & 32.44 \\
LLaVA-Video-7B       & 29.60 & 25.67 & 24.59 & 39.32 \\
MiMo-VL-7B           & 44.53 & 40.47 & 40.83 & 52.46 \\
Qwen2.5-VL-72B       & \cellcolor{best2}47.50 & 35.59 & 41.05 & \cellcolor{best2}64.48 \\
Qwen2.5-VL-7B        & 35.23 & 38.90 & 27.32 & 44.15 \\
Qwen3-VL-4B & 42.17 & 27.40 & 48.96 & 42.09 \\
\midrule
\textbf{VepAgent-4B (Ours)}   & \cellcolor{best}\textbf{70.20} & \cellcolor{best}\textbf{83.46} & \cellcolor{best}\textbf{54.78} & \cellcolor{best}\textbf{83.57} \\
\bottomrule
\end{tabular}
}
\vspace{-0.2in}
\end{table}

\textbf{Implementation Details}. 

We conduct experiments on the Qwen3-VL-4B-Instruct model. 
In stage one, we select approximately 4,000 training samples constructed from Sec.~\ref{sec:Training} to form the \emph{futurebench-4K} dataset for SFT. 
For subsequent RL optimization, we use approximately 5,000 validated training samples. 
The training was conducted on eight NVIDIA RTX A5000 GPUs (24 GB memory each). 
SFT is performed for 1 epoch with an effective batch size of 16 and a learning rate of $5\times10^{-6}$. 
RL is optimized for 250 steps with 8 rollouts per sample and a learning rate of $1\times10^{-6}$. For component-wise ablation studies, we use $G=4$ unless otherwise specified.
More details are shown in Appendix~\ref{appx:exp_details}.

\subsection{Comparison with Reported Baselines}

\textbf{FutureBench}. 
In Tab.~\ref{tab:futurebench}, we evaluate our method on FutureBench under standard protocols against the MLLMs reported in that table. 
Large-scale models such as Qwen3-VL-30B-A3B~\cite{bai2025qwen3} and Qwen2.5-VL-32B~\cite{bai2025qwen2} remain competitive, yet they primarily rely on retrospective summarization of historical observations. This passive observation often induces language-prior shortcuts, where models project future trajectories from superficial textual co-occurrences rather than causal grounding, limiting their ability to bridge unobserved intermediate state transitions.

VepAgent (4B) raises AVG accuracy by \textbf{14.58} percentage points over the prior strongest reported baseline Qwen3-VL-30B-A3B, from 66.86\% to \textbf{81.44\%}, under tool-augmented RL and the same 32-frame budget. Relative to the same-size Qwen3-VL-4B backbone (59.09\% AVG), this corresponds to a gain of 22.35 percentage points. This gain is consistent with improved causal-transition reasoning and visual-evidence recovery.

\textbf{NEPBench}. 
In Tab.~\ref{tab:nepbench}, we further evaluate on NEPBench.
Models such as Qwen2.5-VL-72B remain competitive on specific subsets, yet they struggle with fine-grained detail ambiguity and key-frame omission in cross-domain scenarios under uniform spatio-temporal sampling. By dynamically integrating diagnostic tools to recover missing transient actions, VepAgent raises Overall accuracy by \textbf{22.70} percentage points over the prior strongest reported baseline Qwen2.5-VL-72B, from 47.50\% to \textbf{70.20\%}. Relative to the same-size Qwen3-VL-4B backbone (42.17\% Overall), the gain is 28.03 percentage points. On the reported subsets, VepAgent also attains the highest accuracy: Charades 83.46\% (next, GLM-4.1V-9B at 45.98\%), VidSitu 54.78\% (next, Kimi-VL-A3B at 53.06\%), and YouCook2 83.57\% (next, 64.48\% for both GLM-4.1V-9B and Qwen2.5-VL-72B), under the same evaluation protocol.

\begin{figure*}[t]
    \centering
    \includegraphics[width=\textwidth]{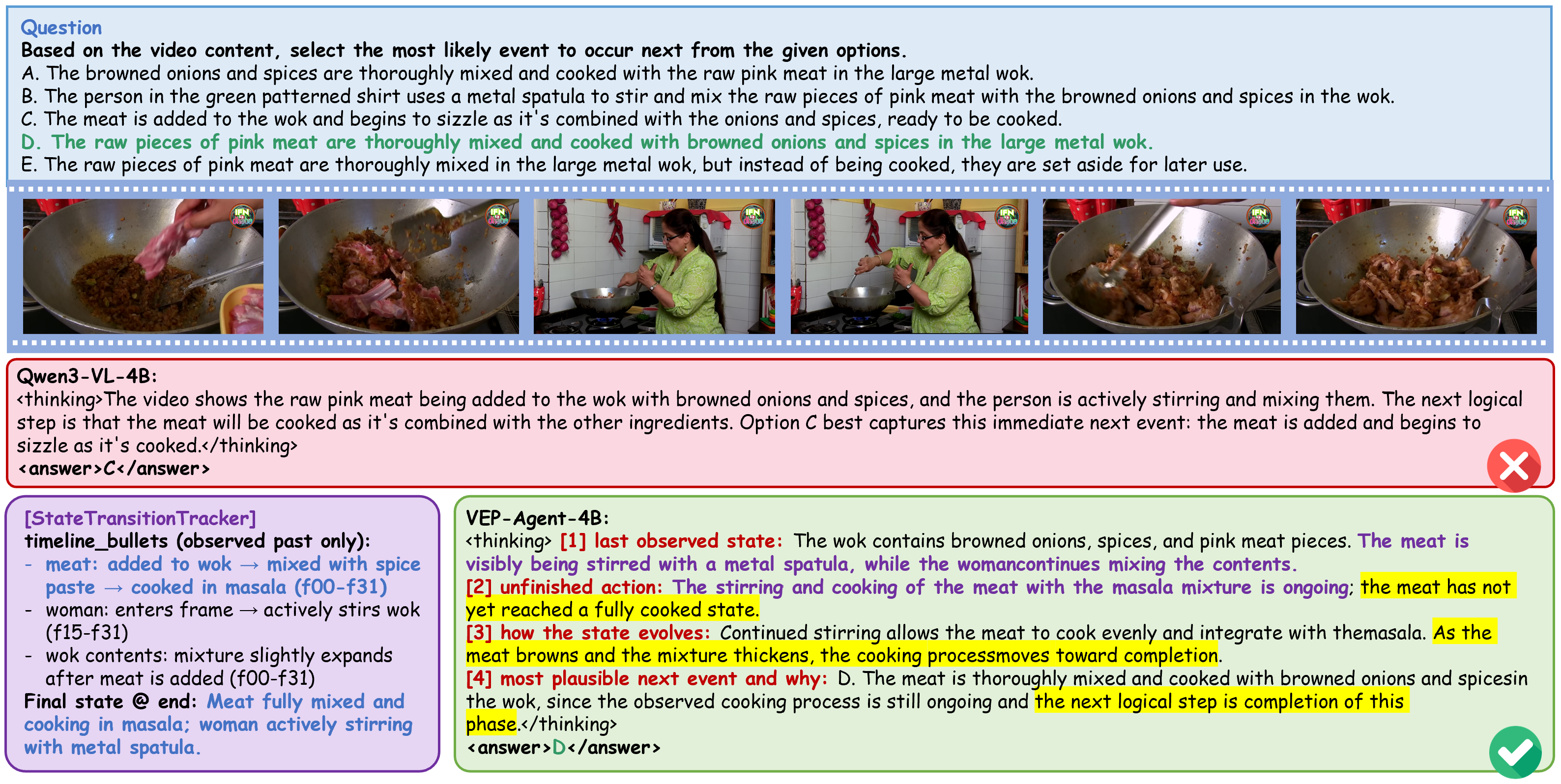}
    \caption{
    \textbf{Case study between Qwen3-VL-4B and VepAgent-4B.}
    The baseline predicts an already observed meat-addition event.
    StateTransitionTracker reconstructs the observed progression from meat addition
    to mixing and active cooking.
    Grounded in evidence, VepAgent identifies the unfinished cooking process,
    reasons over subsequent state evolution, and predicts the correct future event.
    }
    \vspace{-0.2in}
    \label{fig:case_study}
\end{figure*}

\textbf{Qualitative Analysis.}
Beyond the quantitative gains, Fig.~\ref{fig:case_study} illustrates how tool-grounded causal reasoning corrects temporal-state misinterpretation. Although the base model recognizes the overall cooking context, it predicts an event that has already occurred. By explicitly anchoring the terminal observed state, VepAgent reasons forward from the unfinished process rather than replaying the observed action. More qualitative examples are provided in Appendix~\ref{appx:case_study}.

\subsection{Ablation Studies}

\begin{table*}[t]
\centering
\caption{Ablation studies (\%) on FutureBench. Component-wise ablations are conducted with $G=4$.\colorbox{best}{best} and \colorbox{best2}{second best} results are highlighted.}
\vspace{-0.5em}
\label{table: Ablation}
\resizebox{\textwidth}{!}{
\begin{tabular}{l|ccc|l|ccc}
\toprule
\multicolumn{1}{c|}{Method} & \multicolumn{1}{c}{1-Hop $\uparrow$} & \multicolumn{1}{c}{3-Hop $\uparrow$} & \multicolumn{1}{c|}{AVG $\uparrow$} & \multicolumn{1}{c|}{Tools} & \multicolumn{1}{c}{1-Hop $\uparrow$} & \multicolumn{1}{c}{3-Hop $\uparrow$} & \multicolumn{1}{c}{AVG $\uparrow$} \\
\midrule
\emph{(a)} Qwen3-VL-4B & 54.91 & 59.20 & 59.09 & \emph{(e)} w/o. STT & 82.08 & 71.64 & 75.66 \\
\emph{(b)} 4B + SFT only & 77.46 & \cellcolor{best2}\textbf{75.62} & 71.59 & \emph{(f)} w/o. FRE & 82.08 & \cellcolor{best2}\textbf{75.62} & 76.99 \\
\emph{(c)} 4B + SFT + RL & \cellcolor{best2}\textbf{79.77} & 71.14 & \cellcolor{best2}\textbf{75.09} & \emph{(g)} w/o. RFM & \cellcolor{best2}\textbf{82.66} & \cellcolor{best}\textbf{76.12} & \cellcolor{best}\textbf{78.31} \\
\emph{(d)} 4B + SFT + RL + Tools ($G=4$) & \cellcolor{best}\textbf{83.82} & \cellcolor{best}\textbf{76.12} & \cellcolor{best}\textbf{78.79} & \emph{(h)} w/o. VDE & \cellcolor{best}\textbf{83.24} & 75.12 & \cellcolor{best2}\textbf{77.46} \\
\cmidrule(lr){1-4} \cmidrule(lr){5-8}
\multicolumn{1}{c|}{Reward Term} & \multicolumn{1}{c}{1-Hop $\uparrow$} & \multicolumn{1}{c}{3-Hop $\uparrow$} & \multicolumn{1}{c|}{AVG $\uparrow$} & \multicolumn{1}{c|}{Num of Gen.} & \multicolumn{1}{c}{1-Hop $\uparrow$} & \multicolumn{1}{c}{3-Hop $\uparrow$} & \multicolumn{1}{c}{AVG $\uparrow$} \\
\cmidrule(lr){1-4} \cmidrule(lr){5-8}
\emph{(i)} $R$ = $R_{\text{acc}}$ & 78.61 & \cellcolor{best2}\textbf{76.12} & 74.62 & \emph{(m)} \emph{Num = 2} & \cellcolor{best}\textbf{84.39} & \cellcolor{best2}\textbf{77.61} & 78.60 \\
\emph{(j)} $R$ = $R_{\text{acc}}$ + $R_{\text{cau}}$ & \cellcolor{best2}\textbf{82.08} & \cellcolor{best}\textbf{78.11} & \cellcolor{best2}\textbf{76.42} & \emph{(n)} \emph{Num = 4} & \cellcolor{best2}\textbf{83.82} & 76.12 & 78.79 \\
\emph{(k)} $R$ = $R_{\text{acc}}$ + $R_{\text{ant}}$ & 76.88 & \cellcolor{best}\textbf{78.11} & 76.33 & \emph{(o)} \emph{Num = 6} & 82.08 & 76.72 & \cellcolor{best2}\textbf{78.98} \\
\emph{(l)} $R$ = $R_{\text{acc}}$ + $R_{\text{cau}}$ + $R_{\text{ant}}$ (Ours) & \cellcolor{best}\textbf{83.82} & \cellcolor{best2}\textbf{76.12} & \cellcolor{best}\textbf{78.79} & \emph{(p)} \emph{Num = 8} & 82.08 & \cellcolor{best}\textbf{80.10} & \cellcolor{best}\textbf{81.44} \\
\bottomrule
\end{tabular}
}
\vspace{-1em}
\end{table*}

\textbf{Training Stages}.
In rows \emph{(a)-(d)} of Tab.~\ref{table: Ablation}, we compare zero-shot, SFT-only, SFT+RL, and the full model training paradigms, respectively. The zero-shot baseline in row \emph{(a)} yields the lowest AVG accuracy of 59.09\%. SFT-only training in row \emph{(b)} raises AVG accuracy from 59.09\% to 71.59\%. Combining SFT with RL in row \emph{(c)} further raises AVG accuracy to 75.09\%, and the full tool-augmented model in row \emph{(d)} reaches 78.79\%. We attribute this progression to the difficulty of RL in directly navigating the high-dimensional causal reasoning space without prior SFT guidance. These results support our two-stage strategy, where SFT provides a structured reasoning foundation before RL optimization.


\textbf{Tool-Usage}.
In rows \emph{(e)-(h)} of Tab.~\ref{table: Ablation}, we show the effect of removing each specific diagnostic tool under the $G=4$ setting. Relative to the full model in row \emph{(d)} (78.79\% AVG), removing STT in row \emph{(e)}, FRE in row \emph{(f)}, RFM in row \emph{(g)}, and VDE in row \emph{(h)} reduces AVG accuracy to 75.66\%, 76.99\%, 78.31\%, and 77.46\%, respectively, corresponding to absolute drops of 3.13, 1.80, 0.48, and 1.33 percentage points. Removing STT causes the largest drop, while the other tools provide complementary gains for future extrapolation. The full model under this setting achieves the highest AVG score, demonstrating the complementary contributions of the diagnostic tools to VEP.


\textbf{Reward Functions}.
In rows \emph{(i)-(l)} of Tab.~\ref{table: Ablation}, we analyze the effect of different reward configurations in RL training. Using only the accuracy reward \emph{(i)} yields the lowest AVG accuracy of 74.62\%, showing that accuracy and format rewards alone are insufficient to guide causal reasoning. Including the causal gap reward in row \emph{(j)} raises AVG accuracy from 74.62\% to 76.42\%, illustrating that explicit causal-logic bridging provides useful logical cues even without penalizing language shortcuts. Including the anti-prior reward \emph{(k)} similarly raises AVG accuracy from 74.62\% to 76.33\%, confirming that mitigating language-prior shortcuts encourages visual grounding. The full reward in row \emph{(l)} reaches the highest AVG accuracy among these reward settings, raising it from 74.62\% under the accuracy-only reward \emph{(i)} to \textbf{78.79\%}.


\textbf{Number of Generations}.
In rows \emph{(m)-(p)} of Tab.~\ref{table: Ablation}, we adopt different numbers of generations during the RL stage (GRPO). Increasing the number of candidate responses from 2 to 8 raises AVG accuracy from 78.60\% to 81.44\%, although individual metrics show some fluctuations: \emph{Num = 2} yields the highest 1-Hop score (84.39\%), while \emph{Num = 8} (row \emph{(p)}) reaches the highest 3-Hop and AVG scores (\textbf{80.10\%} and \textbf{81.44\%}). We therefore adopt \emph{Num = 8} for our final model.

\section{Conclusion}
We introduced \textbf{VepAgent}, an agentic framework for Video Event Prediction (VEP) that unifies causal-transition reasoning with multimodal tool augmentation. On FutureBench and NEPBench, VepAgent (4B) raises average accuracy by 14.58 and 22.70 percentage points over the prior strongest reported MLLM baselines, from 66.86\% to 81.44\% and from 47.50\% to 70.20\%, respectively, and also improves over the same-size Qwen3-VL-4B backbone under identical protocols. The method comprises three components: (i) the \emph{futurebench-4K} CoT dataset for supervised fine-tuning (SFT), which structures the deduction of unobserved intermediate states; (ii) a diagnostic tool library integrating state tracking, frame retrieval, and region magnification to recover missing spatio-temporal evidence and resolve visual ambiguities; and (iii) a composite reward that jointly optimizes prediction accuracy, causal coherence, and anti-prior robustness, encouraging visual grounding rather than superficial textual similarities. The present evaluation covers FutureBench and NEPBench. Broader video understanding settings remain untested. Future work will expand the diagnostic tool library and apply causal-transition reasoning to those settings.

\subsection*{AI use statement}

Generative AI was used to synthesize CoT reasoning trajectories for constructing training data, and to provide external textual descriptions of visual content via tool-augmented inference. AI tools also assisted with code modification, language editing, and manuscript assessment and revision. It is important to note that AI was not involved in the core research ideas, methodology, or experimental design. The authors take full responsibility for the scientific claims, experimental results, citations, and final manuscript.

\bibliography{iclr2027_conference}
\bibliographystyle{iclr2027_conference}

\newpage

\appendix
\section*{Appendix}
In this appendix, we provide more experimental details, a detailed related work, tool library, and discussions for a comprehensive evaluation and understanding of our method. Detailed contents are as follows:

\setlength{\cftbeforesecskip}{0.5em}
\cftsetindents{section}{0em}{1.8em}
\cftsetindents{subsection}{1em}{2.5em}
\etoctoccontentsline{part}{Appendix}
{
  \etocsettocstyle{}{}
  \localtableofcontents
}

\section{Experiment}
\subsection{More Experimental Details}
\label{appx:exp_details}
\textbf{Datasets.} 
We conduct experiments on FutureBench~\citep{wang2026fostering} and
NEPBench~\citep{lei2020vlep}. FutureBench evaluates future-event prediction under different reasoning depths, including 1-Hop, 2-Hop, 3-Hop, and Interpolation settings. NEPBench contains diverse real-world scenarios from Charades, VidSitu, and
YouCook2. All SFT and RL training trajectories are constructed exclusively from the
training splits to avoid data leakage.

\textbf{Evaluation metrics.}
We follow the evaluation protocols of the corresponding benchmarks and report
prediction accuracy. For FutureBench, we report 1-Hop, 2-Hop, 3-Hop, Interp., and their overall average (AVG). For NEPBench, we report the overall accuracy together with the results on Charades, VidSitu, and YouCook2.

\textbf{More Hyperparameter Configurations.}
We use Qwen3-VL-4B-Instruct as the backbone. SFT is performed on approximately 4K constructed CoT trajectories for one epoch with an effective batch size of 16 and a learning rate of $5\times10^{-6}$. For RL, we use approximately 5K validated trajectories and optimize the model with GRPO for 250 steps with a learning rate of $1\times10^{-6}$ and a group size of $G=8$. We set the KL coefficient $\beta$ to 0.04 and the clipping ratios $\epsilon_{\mathrm{low}}$ and $\epsilon_{\mathrm{high}}$ to 0.2 and 0.3. All experiments are conducted on eight NVIDIA RTX A5000 GPUs with 24 GB memory.

\subsection{Reward Implementation Details}
\label{appx:reward_details}

\textbf{Causal Gap Reward.}
For the causal gap reward, we separately evaluate the terminal observed state and the causal transition toward the future event. Let $s_1$ denote the generated terminal-state description in reasoning step [1], and let $s_{34}$ denote the concatenation of reasoning steps [3] and [4]. We compare them with the corresponding reference terminal state $r_1$ and causal-transition reference $r_4$, respectively. Semantic similarity is computed using MiniLM sentence embeddings with cosine similarity. The raw causal consistency score is defined as
\begin{equation}
g_{\mathrm{raw}}
=
\frac{
0.45\,\mathrm{sim}(s_1,r_1)
+
0.55\,\mathrm{sim}(s_{34},r_4)
}{
0.45+0.55
}.
\end{equation}
If only one reasoning block is available, we compute the score using the available block with its weight renormalized. If both blocks are missing, the complete \texttt{<thinking>} content is used as a fallback.

To prevent an incorrect prediction from receiving excessive process reward, the causal reward is further conditioned on prediction correctness:
\begin{equation}
R_{\mathrm{cau}} =
\begin{cases}
\min(1,\,0.20+0.80g_{\mathrm{raw}}), & R_{\mathrm{acc}}=1,\\
\min(0.35,\,0.20g_{\mathrm{raw}}), & R_{\mathrm{acc}}=0.
\end{cases}
\end{equation}

\textbf{Anti-Prior Reward.}
For the anti-prior reward, we encode the query and each candidate option using a MiniLM sentence encoder and compute their cosine similarities. Candidate options are ranked according to their textual similarity to the query, and the rank of the correct option is used to compute $R_{\mathrm{ant}}$ as defined in the main paper.

\textbf{Reward Ablations.}
For reward-component ablations, the format reward remains active as the hard structural gate. When removing an individual reward term, the original weights of the remaining terms are retained without renormalization.

\subsection{More Ablation Studies}
\label{appx:ablation_study}

To further validate our design choices, we conduct comprehensive ablation studies on the FutureBench dataset under the same $G{=}4$ setting as the component-wise ablations in the main paper.

\begin{table}[ht]
\centering
\caption{Ablation on tool selection strategies.}
\vspace{-1em}
\label{tab:ab_tool_strategy}
\resizebox{0.55\textwidth}{!}{
\begin{tabular}{l|cccc}
\toprule
\multicolumn{1}{c|}{Strategy} & \multicolumn{1}{c}{1-Hop} & \multicolumn{1}{c}{3-Hop} & \multicolumn{1}{c}{AVG} & \multicolumn{1}{c}{Avg Calls} \\ \midrule
\emph{(a)} No Tools & 79.77 & 71.14 & 75.09 & 0.00 \\
\emph{(b)} All Tools (Static) & 76.30 & 70.65 & 74.91 & 4.00 \\
\emph{(c)} Random Tools & 82.08 & 77.61 & 76.99 & 1.85 \\
\emph{(d)} Dynamic (Ours) & \cellcolor{best}\textbf{83.82} & \cellcolor{best}\textbf{76.12} & \cellcolor{best}\textbf{78.79} & 1.13 \\
\bottomrule
\end{tabular}
}
\vspace{-0.5em}
\end{table}

\textbf{(a) Tool Selection Strategy.}
In Tab.~\ref{tab:ab_tool_strategy}, we compare our dynamic selection against: (a) No Tools, (b) All Tools (statically invoking all 4 tools), and (c) Random Tools. Static tool invocation in (b) yields 74.91\% AVG, slightly below the tool-free baseline (75.09\%), as forcibly injecting all tool outputs can introduce irrelevant evidence that distracts the model from the true causal cues. Random selection in (c) raises AVG accuracy from 75.09\% to 76.99\% over No Tools, but remains below Dynamic and uses more average tool calls (1.85 vs.\ 1.13), reflecting the cost of sometimes retrieving uncorrelated spatial or temporal evidence. Our Dynamic Selection in (d) achieves the highest AVG accuracy (78.79\%) and the best 1-Hop score (83.82\%), with an average call count of 1.13. Random attains a higher 3-Hop score (77.61\% vs.\ 76.12\%), yet Dynamic remains strongest on AVG. This demonstrates that the RL agent learns to retrieve evidence selectively when the confidence of $S_{\text{final}}$ and $A_{\text{ongoing}}$ is low. Relative to Random, Dynamic raises AVG accuracy from 76.99\% to 78.79\% while using fewer tool calls (1.13 vs.\ 1.85).

\begin{table}[ht]
\centering
\caption{Ablation on tool synergy between temporal and spatial dimensions.}
\vspace{-1em}
\label{tab:ab_tool_synergy}
\resizebox{0.55\textwidth}{!}{
\begin{tabular}{l|ccc}
\toprule
\multicolumn{1}{c|}{Tool Configuration} & \multicolumn{1}{c}{1-Hop} & \multicolumn{1}{c}{3-Hop} & \multicolumn{1}{c}{AVG} \\ \midrule
\emph{(a)} Temporal Only (STT + FRE) & 82.08 & 75.62 & 77.84 \\
\emph{(b)} Spatial Only (RFM + VDE) & 80.92 & 71.64 & 75.38 \\
\emph{(c)} All (Ours) & \cellcolor{best}\textbf{83.82} & \cellcolor{best}\textbf{76.12} & \cellcolor{best}\textbf{78.79} \\
\bottomrule
\end{tabular}
}
\vspace{-0.5em}
\end{table}

\textbf{(b) Synergy of Spatio-Temporal Tools.}
In Tab.~\ref{tab:ab_tool_synergy}, We evaluate the agent using only temporal or spatial tools. Relying solely on spatial tools (RFM, VDE) in (b) allows the model to verify local details but fails to capture chronological dynamics, yielding the weakest 3-Hop accuracy (71.64\%) and the lowest AVG (75.38\%). Conversely, temporal tools (STT, FRE) in (a) clarify action sequences and reach 77.84\% AVG, but still fall short of the full model on 1-Hop and AVG. Combining both dimensions in (c) achieves the best overall results (83.82\% / 76.12\% / 78.79\% on 1-Hop / 3-Hop / AVG), indicating that temporal tracking and spatial magnification are complementary for resolving visual-evidence omissions in VEP. Specifically, temporal tools narrow down the critical time windows, which subsequently allows spatial tools to perform precise verifications on the exact frames, forming a cohesive evidence-gathering pipeline.

\subsection{More Case Studies}
\label{appx:case_study}

To further illustrate how diagnostic tools support causal-transition reasoning, we provide two additional qualitative examples. These cases highlight two representative failure modes in VEP: missing temporal evidence and fine-grained state ambiguity. In both cases, the baseline tends to repeat an event that has already occurred, whereas VepAgent uses additional evidence from the observed video to better determine the terminal state and reason toward the subsequent event.

\begin{figure}[t]
    \centering
    \includegraphics[width=\textwidth]{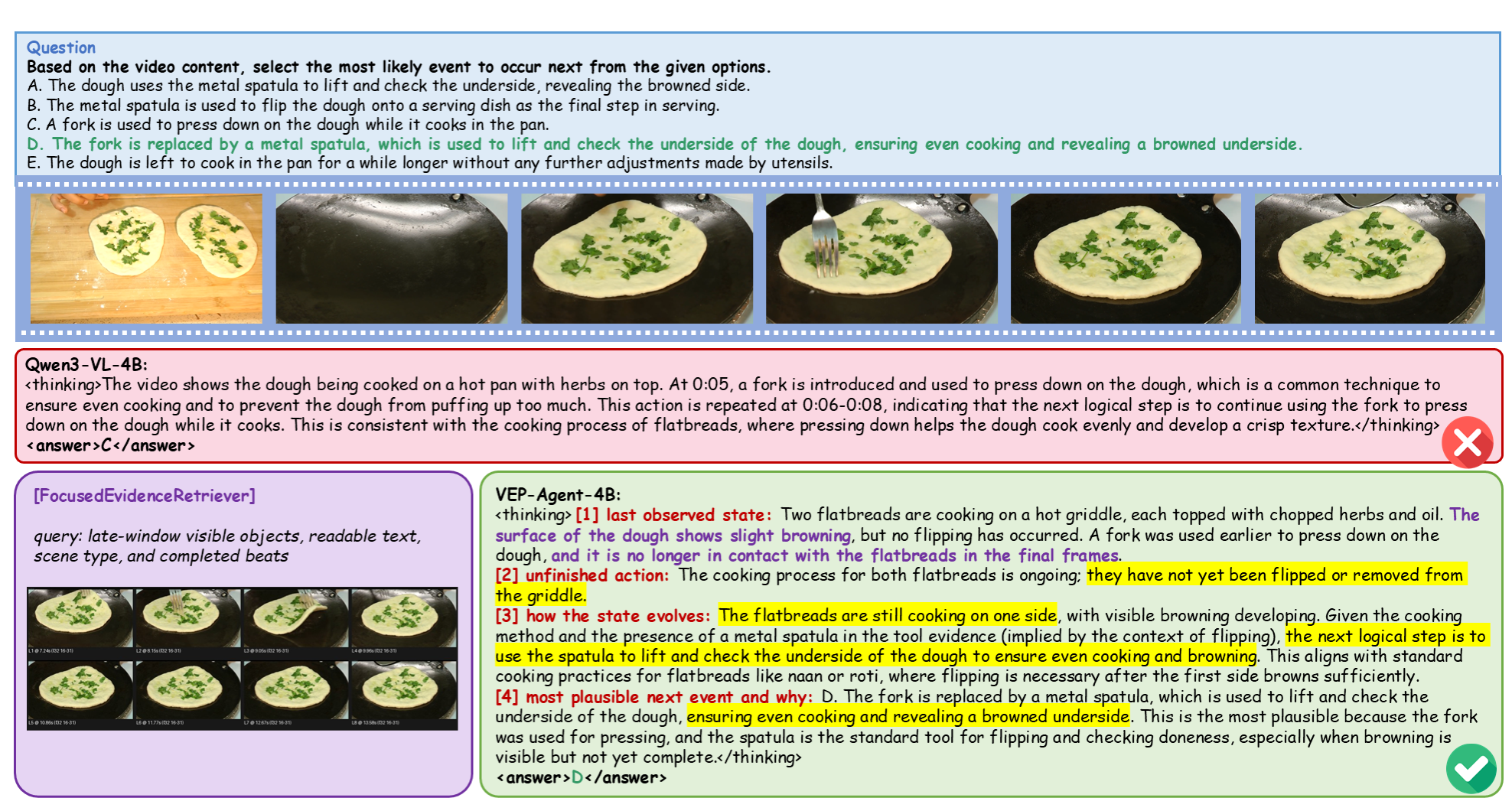}
    \caption{
    \textbf{Additional case study with FocusedEvidenceRetriever.}
    The baseline repeatedly predicts the earlier fork-pressing action.
    FRE retrieves dense late-window visual evidence showing that the
    fork interaction has ended while the flatbread continues browning.
    VepAgent uses this temporal evidence to reason toward the subsequent
    cooking transition.
    }
    \label{fig:case_fre}
\end{figure}

\textbf{Recovering Missing Temporal Evidence.}
As shown in Fig.~\ref{fig:case_fre}, sparse observation can make the model overemphasize an earlier salient action and incorrectly treat it as the future event. By retrieving denser evidence from the relevant late temporal window, FRE reveals that the fork interaction has already finished while the cooking process remains ongoing. This additional evidence helps VepAgent distinguish the completed action from the unfinished process and therefore reason toward the next plausible cooking event.

\begin{figure}[t]
    \centering
    \includegraphics[width=\textwidth]{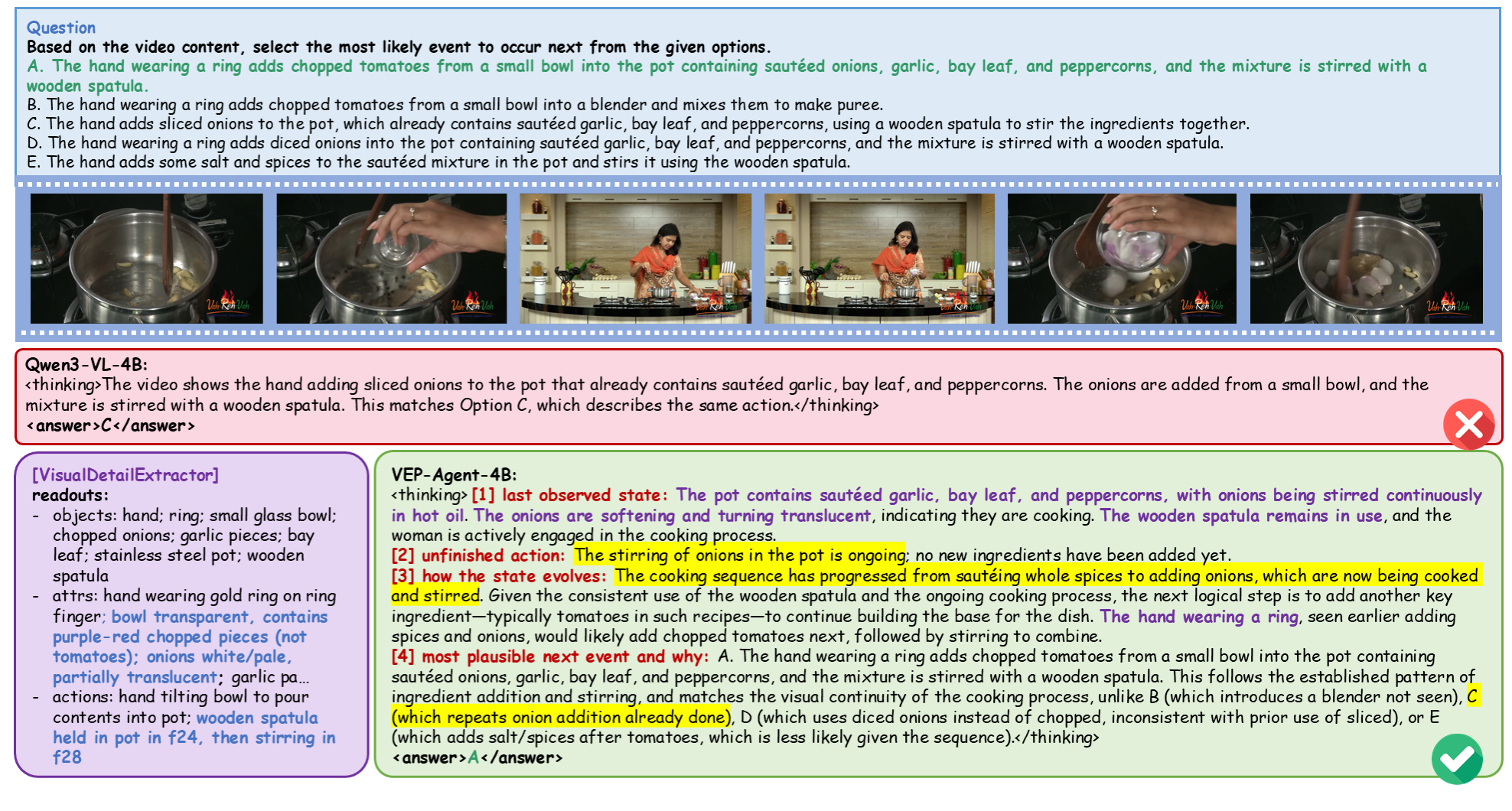}
    \caption{
    \textbf{Additional case study with VisualDetailExtractor.}
    The baseline repeats the already observed onion-addition event.
    VDE verifies fine-grained visual states, including partially translucent
    onions being stirred in the pot.
    VepAgent therefore recognizes that onion addition has completed and
    reasons toward the next ingredient stage.
    }
    \label{fig:case_vde}
\end{figure}

\textbf{Resolving Fine-Grained State Ambiguity.}
Fig.~\ref{fig:case_vde} demonstrates a complementary failure mode in which the overall scene is correctly recognized but the completion state of an action is misinterpreted. VDE extracts explicit visual facts from the final observed frames, allowing VepAgent to verify that the onion-addition stage has already completed and that the cooking process has progressed further. Grounded in this terminal-state evidence, the model avoids repeating the observed event and instead predicts the subsequent ingredient-related transition.

\section{Detailed Related Work}
\label{appx:related_work}
This section extends the concise related work in the main paper with a broader review of both directly related and complementary research directions. We first discuss video event prediction, multimodal reasoning with reinforcement learning, and tool-augmented agentic systems, which are most closely related to VepAgent. We then review adjacent areas, including embodied vision-language-action reasoning, spatio-temporal and geometric perception, human-centric interaction understanding, cross-modal representation learning, and domain-specific multimodal applications. Although these adjacent areas address different tasks, they provide complementary perspectives on active perception, evidence recovery, multimodal alignment, and robust reasoning under incomplete observations.

\subsection{Video Event Prediction and Temporal Event Reasoning}
Video Event Prediction (VEP) is closely related to script event prediction, action anticipation, and temporal video understanding. Early work on script event induction models dependencies among events in textual narratives to infer plausible subsequent events~\citep{granroth2016mcnc}. In the visual domain, action anticipation extends this forward-looking objective to videos by predicting future actions from partial observations~\citep{girdhar2021antvc}. Large-scale datasets such as EPIC-Kitchens provide structured annotations of daily activities and support the study of temporal action evolution~\citep{damen2018epic}. VidSitu further represents complex real-world videos with semantic roles and structured event annotations~\citep{sadhu2021vidsitu}.

More recent studies move beyond action-level anticipation toward explicit future-event reasoning. VLEP formulates next-event prediction from partially observed videos and evaluates whether models can infer plausible subsequent events~\citep{lei2020vlep}. VidEvent studies structured video events for action-centric event prediction~\citep{liang2025videvent}. EventFormer models hierarchical dependencies among events to capture structured temporal evolution~\citep{su2025eventformer}. FutureBench evaluates future-event reasoning across different temporal reasoning depths, providing a benchmark for modern multimodal models~\citep{wang2026fostering}. Video-CoE introduces a Chain-of-Events paradigm that explicitly organizes observed temporal events to strengthen the connection between video history and future outcomes~\citep{su2026video}. Human-centric open-future task discovery further investigates how unseen future-oriented tasks can be discovered and organized beyond a predefined label space~\citep{song2026human}.

Despite these advances, most existing approaches primarily improve the representation and organization of observed temporal history. This can remain insufficient when the terminal observed state is ambiguous, an ongoing action has not yet completed, or critical evidence is missed by sparse visual sampling. VepAgent instead focuses explicitly on the transition from the terminal observed state and unfinished action toward the future event. It further equips the model with diagnostic tools that recover missing temporal evidence and verify fine-grained visual states before future-event prediction.

\subsection{Multimodal LLM Reasoning and Reinforcement Learning}
Recent progress in large language models has shown that reinforcement-learning-based post-training can substantially strengthen structured reasoning. OpenAI-o1 demonstrates strong deliberative reasoning through reinforcement-learning-based training~\citep{jaech2024openaio1}. DeepSeek-R1 further advances large-scale reasoning-oriented post-training and demonstrates the effectiveness of reinforcement learning for eliciting complex reasoning behaviors~\citep{deepseekr1}. Similar paradigms have subsequently been extended to multimodal reasoning. LMM-R1 applies reasoning-oriented learning to mathematical visual question answering~\citep{peng2025lmmr1}. Reinforcement learning has also been explored for visual segmentation and grounding, showing its applicability beyond text-centric reasoning~\citep{liu2025seg}. Video-R1 extends reinforcement-learning-based reasoning to video multimodal large language models~\citep{feng2025videor1}. OneThinker further studies unified reasoning across image and video understanding tasks~\citep{feng2026onethinker}.

A related line of research investigates how reasoning policies themselves can be improved through better trajectories and reward signals. EchoRL enhances reinforcement learning by reusing informative rollout signals during optimization~\citep{bi2026echorl}. ReflectRL learns from golden negative trajectories through a reflective-to-direct reasoning procedure~\citep{bi2026reflectrllearninggoldennegative}. Rubric-based reward and guidance have also been introduced to encourage more effective exploration in multi-domain reasoning~\citep{bi2025reward}.

Beyond direct policy optimization, recent studies improve reasoning through data refinement, data selection, test-time decoding, and internal trajectory analysis. RefineX learns to refine pre-training data through expert-guided programs~\citep{bi2025refinex}. PRISM studies training-free multimodal data selection to identify informative examples for multimodal learning~\citep{Bi2025PRISMSI}. PromptCD improves test-time behavior through polarity-prompt contrastive decoding~\citep{bi2026promptcd}. Layerwise trajectory analysis further investigates the internal geometry of reasoning and model self-evaluation across network layers~\citep{bi2026the}.

Reinforcement learning has also been increasingly explored for video-specific multimodal tasks. CaC improves video reward modeling through hierarchical spatio-temporal concentrating~\citep{wang2026cac}. FactGuard applies agentic reinforcement learning to video misinformation detection, illustrating the potential of RL for complex video-level reasoning~\citep{li2026factguard}.

Nevertheless, stronger reasoning performance does not necessarily imply stronger utilization of visual evidence. Studies of video reasoning models show that models with enhanced reasoning capabilities may still under-utilize visual information~\citep{chen2025seedbenchr1}. Multimodal models can also rely heavily on language-derived priors when solving temporally complex problems~\citep{chen2025longvilar1}. Such behavior is particularly problematic for VEP, where subtle differences in the terminal observed state can lead to substantially different future trajectories. VepAgent therefore combines reasoning-oriented reinforcement learning with explicit causal-gap and anti-prior objectives, encouraging future-event predictions to remain grounded in the observed video rather than textual shortcuts.

\subsection{Tool-Augmented Agentic Systems}
Another closely related direction explores the use of external tools to acquire, transform, or verify visual evidence during multimodal reasoning. FAST incorporates visual evidence directly into intermediate reasoning steps for image understanding~\citep{sun2025fast}. MVoT constructs multimodal chains of thought through interleaved visual operations~\citep{li2025MVoT}. LLaVA-Plus investigates how multimodal assistants can learn to invoke external tools for solving complex visual tasks~\citep{liu2023llavapluslearningusetools}. Visual Program Distillation transfers tool-use skills derived from executable programs into vision-language models~\citep{hu2024visual}.

More recent approaches integrate tool use more tightly with the reasoning process. TACO trains multimodal action models with interleaved reasoning and tool invocation~\citep{ma2024taco}. PyVision enables multimodal large language models to dynamically generate and execute visual-processing programs during inference~\citep{zhao2025pyvision}. VITAL further introduces tool-augmented reinforcement learning for long-video reasoning, allowing models to acquire visual evidence on demand~\citep{zhang2025thinking}.

Tool-augmented reinforcement learning has also been extended to task-specific visual domains. Video-STAR reinforces open-vocabulary action recognition through agentic tool use~\citep{yuan2025video}. IndusAgent applies tool-augmented reinforcement learning to open-vocabulary industrial anomaly detection~\citep{tan2026indusagent}. AutoDesign studies meta-harness optimization for long-horizon agentic design and demonstrates the benefit of optimizing external interaction strategies~\citep{luo2026autodesign}. Collectively, these studies suggest that multimodal agents can benefit from actively acquiring task-relevant evidence instead of relying exclusively on a fixed visual representation.

Most existing tool-augmented frameworks, however, are designed for general visual reasoning, recognition, or long-video understanding rather than the specific causal requirements of future-event prediction. In VEP, tool use must help determine what has already happened, what action remains unfinished, and which visual evidence is necessary for predicting the next event. VepAgent therefore introduces a VEP-oriented diagnostic tool library targeting chronological state ambiguity, missing transient actions, local-region ambiguity, and overlooked fine-grained visual facts.

\subsection{Vision-Language-Action Models and Embodied Reasoning}
Vision-Language-Action (VLA) models provide another relevant perspective by integrating visual perception, language reasoning, and action generation within a unified embodied decision-making framework. A recent survey summarizes the development and remaining challenges of pure VLA models~\citep{zhang2025pure}. Reasoning-VLA studies efficient and spatially guided general reasoning for vision-language-action models in autonomous driving~\citep{zhangreasoning}. CoC-VLA introduces chain-of-causality modeling to improve the interpretability of visual-language-action reasoning for autonomous driving~\citep{NEURIPS2025_66d09284}. AutoDrive-R2 further encourages reasoning and self-reflection capabilities in VLA models through dedicated training objectives~\citep{yuan2025autodrive}.

Recent embodied models also explore more general forms of unified perception, reasoning, and action. Pelican-Unify studies unified embodied intelligence encompassing understanding, reasoning, imagination, and action~\citep{zhang2026pelican}. Pelican-VLA emphasizes the importance of attending to relevant visual evidence before action generation and reports improved generalization~\citep{ding2026pelican}. Robotic manipulation has also been formulated through direct vision-to-geometry mapping, providing an alternative to purely language-centric interfaces for embodied interaction~\citep{song2026robotic}.

End-to-end autonomous driving systems further illustrate the importance of structured visual perception and decision making under dynamic observations. MapExpert studies online high-definition map construction with sparse map-element experts~\citep{zhang2025mapexpert}. EHSS develops an efficient hybrid-supervised stereo matching network for transportation scenes~\citep{zhang2023ehss}. AT-Drive exploits adversarial transfer to improve end-to-end autonomous driving~\citep{zhang2025drive}. ADDI simplifies end-to-end driving through distinct experts and implicit interactions~\citep{zhang2025addi}. DiffusionDriveV2 combines truncated diffusion modeling with reinforcement learning for end-to-end driving~\citep{zou2025diffusiondrivev2}.

Although these approaches primarily predict control actions rather than future semantic events, they share with VEP the need to reason from incomplete observations toward a future state. VepAgent addresses a complementary problem: instead of generating embodied controls, it predicts the next event by explicitly bridging unobserved causal transitions and selectively acquiring additional evidence when the observed state is insufficient.

\subsection{Spatio-Temporal Perception and Geometric Evidence}
Although VepAgent does not explicitly reconstruct 3D geometry, advances in geometric perception provide a complementary perspective on recovering reliable visual evidence from incomplete, ambiguous, or occluded observations. Multi-view stereo, in particular, has extensively studied how structural priors can improve visual inference when direct appearance cues are insufficient.

SD-MVS introduces segmentation-driven deformation for improving multi-view stereo reconstruction~\citep{yuan2024sd}. TSAR-MVS further incorporates textureless-aware segmentation and correlative refinement to address regions with weak visual textures~\citep{yuan2024tsar}. MSP-MVS leverages multi-granularity segmentation priors to enhance multi-view geometric estimation~\citep{yuan2025msp}. DVP-MVS integrates depth-edge and visibility priors to improve reconstruction reliability~\citep{yuan2025dvp}. DVP-MVS++ further harmonizes depth, normal, edge, and visibility cues within multi-view stereo~\citep{yuan2025dvpplusplus}. SED-MVS combines segmentation-driven deformation with depth restoration and occlusion constraints~\citep{yuan2025sed}. MonoMVSNet incorporates monocular priors to guide multi-view stereo estimation~\citep{monomvsnet}. MVSMamba explores state-space modeling for capturing long-range dependencies in multi-view stereo~\citep{mvsmamba}.

Beyond reconstruction, geometry-aware modeling has also been explored in visual editing, generation, and dynamic-scene understanding. Geometry-guided reinforcement learning improves multi-view consistency in 3D scene editing~\citep{wang2026geometry}. Dense geometry estimation can also be transferred from representations originally developed for editing-oriented tasks~\citep{wang2026edit}. Layout2Scene synthesizes 3D scenes from semantic layouts by modeling geometric and appearance information~\citep{chen2025layout2scene}. Graph2Scene generates indoor scenes through interaction-aware scene graphs~\citep{chen2025graph2scene}. Adaptive test-time scaling has been explored for improving controllable image editing~\citep{qu2026scale}. STDR studies spatio-temporal decoupling for efficient rendering of dynamic scenes~\citep{li2025stdr}. GradiSeg improves boundary precision for segmentation in 3D Gaussian representations~\citep{li2024gradiseg}.

These studies differ substantially from VEP in task formulation, but they collectively emphasize several principles that are also important for future-event reasoning: exploiting auxiliary visual cues, handling occlusion, recovering fine-grained structure, and avoiding decisions based solely on incomplete direct observations. VepAgent follows a related motivation at the video-reasoning level. Rather than reconstructing full 3D geometry, it selectively retrieves temporal evidence and magnifies task-relevant regions to verify the terminal observed state and support causal-transition reasoning.

\subsection{Human-Centric Interaction and Action Understanding}
Human-centric interaction understanding provides another complementary line of research because future events are frequently determined by how people interact with objects and how such interactions evolve over time. Synergistic prompting has been studied for human-object interaction detection to improve interaction-aware visual reasoning~\citep{luo2025synergistic}. InstructHOI introduces context-aware instructions for multimodal human-object interaction reasoning~\citep{luo2026instructhoi}. Open-vocabulary HOI comprehension has further explored whether models can reason about missing visual contexts through imagination and generation~\citep{yuan2026if}.

Spatial reasoning is also essential for understanding fine-grained interactions. SIRI-Bench evaluates the spatial intelligence of vision-language models through complex reasoning tasks~\citep{song2025siri}. Video-STAR studies open-vocabulary action recognition with tool-augmented reinforcement learning, allowing models to actively acquire additional evidence for recognizing actions~\citep{yuan2025video}.

These approaches emphasize accurate grounding of people, objects, actions, and their spatial relationships. VepAgent shares the need for such fine-grained state verification, but focuses on a different temporal objective. Specifically, it reasons from an observed interaction that may still be incomplete toward the unobserved transition that determines the subsequent event.

\subsection{Cross-Modal Retrieval, Representation Learning, and Multimodal Fusion}
Another complementary direction concerns cross-modal alignment and representation learning. VEP requires a model to align visual states with textual queries and candidate future events, making reliable multimodal representation particularly important when visual evidence is ambiguous.

ConeSep studies robust noise-unlearning for composed image retrieval~\citep{ConeSep}. TEMA improves multi-modification composed image retrieval by anchoring visual representations to the source image while following textual modifications~\citep{TEMA}. ReTrack proposes evidence-driven dual-stream directional anchor calibration for composed video retrieval~\citep{ReTrack}. These retrieval-oriented methods highlight the importance of maintaining reliable visual-textual alignment when multiple semantic factors must be jointly considered.

Related representation-learning methods investigate robustness under incomplete or heterogeneous views. SEC-LSRM improves contrastive multi-view remote sensing clustering through long-short range information mining~\citep{SEC-LSRM}. PAGC addresses multi-view attribute-missing graph clustering through prototype-driven representation learning~\citep{PAGC}. OmniMamba explores state-space models for efficient multimodal understanding and generation~\citep{zou2025omnimamba}. LongCat-Next lexicalizes different modalities into discrete tokens for unified multimodal modeling~\citep{team2026longcat}.

These studies address objectives different from VEP, but they provide useful perspectives on maintaining semantic alignment and robust representations under noisy, missing, or heterogeneous multimodal evidence. In VepAgent, cross-modal alignment is not the final retrieval objective; instead, it supports the causal reasoning process that connects observed visual states to candidate future events.

\subsection{Domain-Specific Multimodal Reasoning and Applications}
Beyond general video and vision-language understanding, a broad range of domain-specific systems study structured multimodal perception and reasoning under challenging observations. These works provide complementary evidence that domain-specific priors, multimodal fusion, and adaptive inference can improve robustness when inputs are incomplete, noisy, or heterogeneous.

In remote sensing, recent work has surveyed the development of spatio-temporal vision-language models and their applications to large-scale Earth observation~\citep{liu2025remoteVLM}. Decoupled prompt learning has also been introduced for remote sensing change captioning, where models must identify and describe meaningful changes across visual observations~\citep{liu2023decoupling}. In biomedical imaging, unsupervised multi-scale registration has been developed to support multimodal microscopy analysis~\citep{wu2026m}. EMPOWER applies evolutionary medical prompt optimization with reinforcement learning to improve task-specific multimodal prompting~\citep{chen2026empower}.

Related ideas also appear outside conventional vision-language tasks. ZeroDiff investigates zero-shot time-series reconstruction through diffusion with informed priors~\citep{pmlr-v306-fan26d}. Simulated lifting provides a deterministic strategy for escaping local minima in non-convex matrix sensing~\citep{shen2026escaping}. Although these problems differ substantially from video event prediction, they similarly emphasize robust inference from incomplete observations and the use of structured priors to reduce ambiguity.

These broader applications reinforce a general motivation shared by VepAgent: reliable reasoning often requires more than directly processing the initially observed input. Instead, the model may benefit from task-specific mechanisms that recover missing information, exploit complementary evidence, and constrain inference using structured knowledge.

\subsection{Summary}
Overall, the reviewed literature can be organized into two complementary groups. The first group is directly related to VepAgent and includes future-event prediction, multimodal reasoning with reinforcement learning, and tool-augmented agentic systems. These studies motivate the importance of structured temporal reasoning, visually grounded policy optimization, and active evidence acquisition. However, existing approaches do not explicitly integrate these capabilities around the specific causal transition from a terminal observed video state to an unobserved future event.

The second group covers adjacent areas including embodied reasoning, geometric perception, human-centric interaction understanding, cross-modal representation learning, and domain-specific multimodal systems. Although these tasks differ from VEP, they provide complementary insights into state verification, spatial reasoning, evidence recovery, multimodal alignment, and robust inference under incomplete observations.

Building on both lines of research, VepAgent focuses specifically on the transition from partially observed video evidence to plausible future events. It combines causal-transition reasoning with diagnostic tool augmentation and anti-prior reinforcement learning, enabling the model to identify the terminal observed state, verify unfinished actions, recover missing evidence when necessary, and reason toward the subsequent event with stronger visual grounding.

\section{Tool Library Details}
\label{appx:tool_library}
This section provides a detailed specification of the multimodal diagnostic tools integrated into our VepAgent framework. These tools are designed to resolve visual-evidence omissions and spatio-temporal ambiguities during Video Event Prediction (VEP) and are categorized into chronological state tracking, temporal evidence retrieval, spatial magnification, and fine-grained fact extraction.

\vspace{-0.1in}
\begin{table}[ht]
\centering
\caption{Implementation details of the diagnostic tool library.}
\vspace{0.1in}
\label{tab:tool_details}
\resizebox{\textwidth}{!}{
\begin{tabular}{cccc}
\toprule
Tool & Input & Output & Failure Handling \\
\midrule
State Transition Tracker (STT) & observed video & textual event chain \& final state & fallback to global video caption \\
Focused Evidence Retriever (FRE) & video \& temporal window & dense frame sequence (contact sheet) & ignored if temporal window is invalid \\
Region Focus Magnifier (RFM) & video \& predicted spatial boxes & magnified image patch & ignored if boxes are invalid \\
Visual Detail Extractor (VDE) & final frames \& scene query & textual fine-grained facts & fallback to original query \\
\bottomrule
\end{tabular}}
\end{table}

\textbf{Tool Execution Protocol.}
For each input, the policy performs a single tool-selection step and selects a subset of diagnostic tools $C \subseteq \mathcal{T}$. The selected tools are executed once, and their returned evidence is incorporated into the reasoning context before the final causal prediction. No additional tool-selection round is performed after receiving the tool outputs. All diagnostic tools operate exclusively on the observed portion of the video and never access future frames.

\textbf{Implementation Backends.}
FRE and RFM are implemented as lightweight visual-processing operators using OpenCV and Pillow. STT and VDE use a DashScope-hosted VLM to respectively construct textual state-transition trajectories and extract fine-grained visible facts.

\subsection{State Transition Tracker}
To resolve ambiguities in action order and state changes, our framework incorporates a \textbf{State Transition Tracker (STT)}. Its primary function is to chronologically re-analyze the observed video and structure the historical event chain into a clear textual sequence. This tool is dynamically invoked when the initial observation suggests conflicting state transitions (e.g., whether the hand transitioned from empty to holding, or vice versa). By explicitly mapping out the sequence of past actions and identifying the exact final observed state (e.g., ``The person is still holding the cup above the table''), the agent can accurately anchor its causal-transition reasoning. This capability is critical for mitigating state/sequence misreading and directly establishes the foundational premise from which the future event is extrapolated.

\subsection{Focused Evidence Retriever}
To address the omission of transient actions caused by uniform frame sampling, we introduce the \textbf{Focused Evidence Retriever (FRE)}. This tool is specifically designed to capture fleeting actions that occur between sparsely sampled frames (e.g., missing the exact moment a cup is placed on a table). When the agent identifies a critical temporal gap where an ongoing action is suspected but unobserved, the FRE performs dense frame extraction within a specified temporal window. These densely sampled frames are then stitched into a contact sheet and fed back into the agent as augmented visual evidence. By recovering these missing spatio-temporal cues, the FRE ensures that crucial intermediate state transitions are not overlooked, enabling robust causal bridging.

\subsection{Region Focus Magnifier}
To resolve local detail ambiguity, our framework employs a \textbf{Region Focus Magnifier (RFM)}. This tool is dynamically invoked when the global view is insufficient to conclusively verify micro-state changes, such as hand-object contact or the precise orientation of a tool. By cropping and magnifying key spatial regions centered on specific human-object interactions, the RFM allows the agent to perform a focused inspection of critical details. This capability is indispensable for accurately determining whether an ongoing action has been completed or is still in progress (e.g., verifying if a hand has truly released a cup or if a door handle has been turned). By magnifying these subtle cues, the RFM directly mitigates failures arising from local detail blur.

\subsection{Visual Detail Extractor}
To mitigate overlooked fine-grained visual facts, we leverage the \textbf{Visual Detail Extractor (VDE)}. This component translates specific, easily ignored visual details from the final observed frames into explicit textual descriptions. When the agent needs to verify specific physical conditions to formulate a causal hypothesis, the VDE provides grounded factual statements (e.g., ``The person's right hand is still holding the cup,'' or ``The cup is moving downward toward the table''). By explicitly textualizing these verifiable visual facts without predicting the future, the VDE anchors the agent's reasoning in concrete evidence. This strict factual grounding prevents the model from taking language-prior shortcuts, forcing it to base its causal gap bridging on actual visual states rather than superficial textual similarities between the query and options.

\section{Prompts}
\label{appx:prompts}

This section provides the prompt templates used in our agentic reasoning pipeline. As described in the main paper, VepAgent follows a two-round interaction pattern. In the first round, the model analyzes the current visual evidence and determines whether additional diagnostic evidence is needed, selecting an appropriate subset of tools when necessary. The selected tools are executed once on the observed portion of the video, and their outputs are returned to the model. In the second round, the model integrates the original observations with the retrieved evidence and performs structured causal-transition reasoning, identifying the terminal observed state, unfinished action, plausible state evolution, and final future-event prediction. The following figures present the corresponding prompt templates used in our framework.

\begin{center}
    \includegraphics[width=\textwidth]{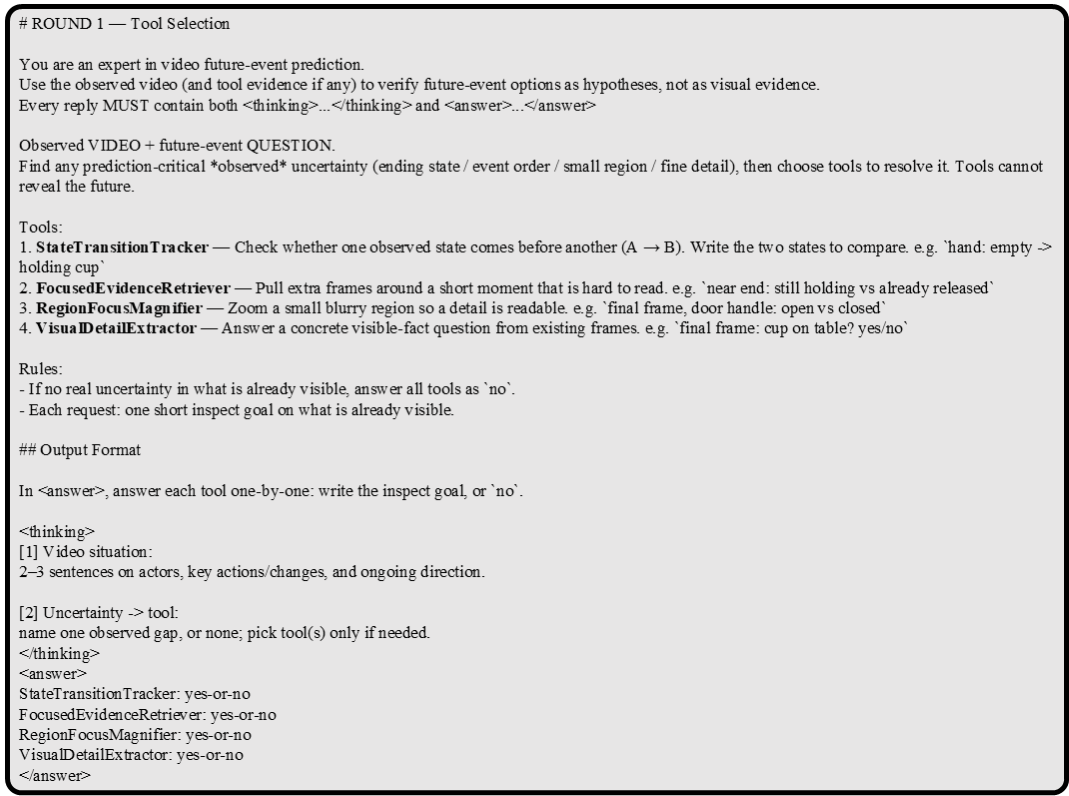}
\end{center}
\vspace{-0.1in}

\begin{center}
    \includegraphics[width=\textwidth]{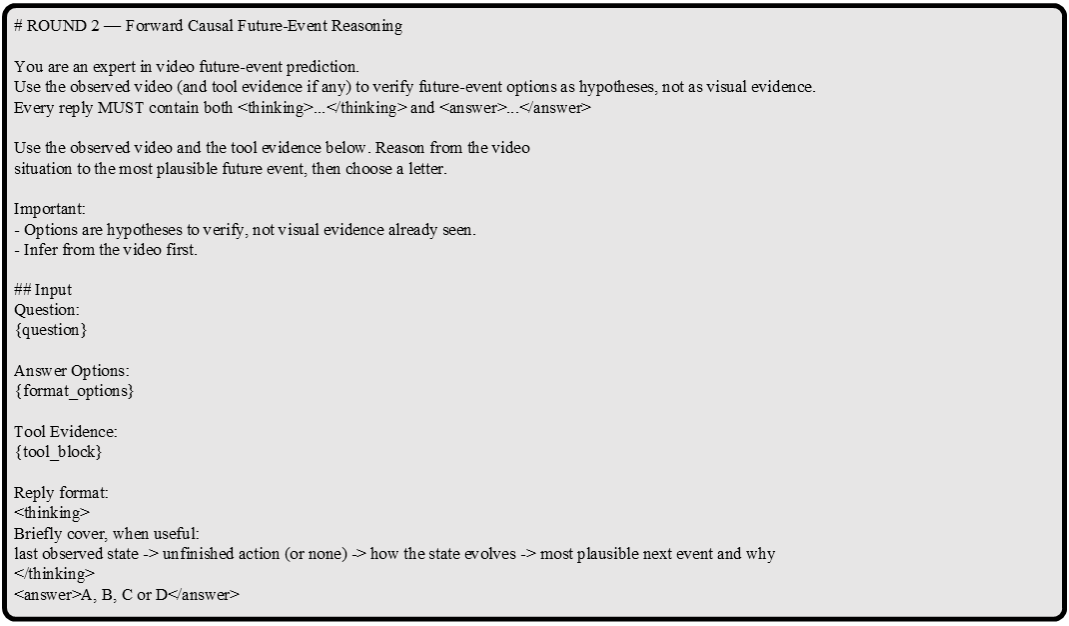}
\end{center}
\vspace{-0.1in}

\end{document}